\documentclass{article}
\usepackage{arxiv}
\usepackage[utf8]{inputenc} 
\usepackage[T1]{fontenc}    
\usepackage{hyperref}       
\usepackage{url}            
\usepackage{booktabs}       
\usepackage{amsfonts}       
\usepackage{nicefrac}       
\usepackage{microtype}      
\usepackage{lipsum}
\usepackage{graphicx}
\usepackage{multicol}
\usepackage{float}
\usepackage{enumitem}
\usepackage{siunitx}
\usepackage[font=small,labelfont=bf]{caption}
\usepackage[backend=biber,style=numeric,sorting=none]{biblatex}
\setlist[itemize]{leftmargin=*}
\newcommand{\beginsupplement}{%
        \setcounter{table}{0}
        \renewcommand{\thetable}{S\arabic{table}}%
        \setcounter{figure}{0}
        \renewcommand{\thefigure}{S\arabic{figure}}%
        \setcounter{section}{0}
     }

\newenvironment{Table}
  {\par\bigskip\noindent\minipage{\columnwidth}\centering}
  {\endminipage\par\bigskip}

\title{Cedille:\\A large autoregressive language model in French}

\author{
  Martin M\"uller\thanks{Authors contributed equally, order is random}\hspace{2cm}Florian Laurent\footnotemark[1] \\ \vspace{0.3cm}\\ Cedille AI\textsuperscript{$1$} \\ \texttt{hello@cedille.ai}
}

\begin{document}
\maketitle

\begin{abstract}
  Scaling up the size and training of autoregressive language models has enabled novel ways of solving Natural Language Processing tasks using zero-shot and few-shot learning.
  While extreme-scale language models such as GPT-3 offer multilingual capabilities, zero-shot learning for languages other than English remain largely unexplored.
  Here, we introduce Cedille, a large open source auto-regressive language model, specifically trained for the French language.
  Our results show that Cedille outperforms existing French language models and is competitive with GPT-3 on a range of French zero-shot benchmarks.
  Furthermore, we provide an in-depth comparison of the toxicity exhibited by these models, showing that Cedille marks an improvement in language model safety thanks to dataset filtering.
\end{abstract}

\vspace{1.5cm}

\begin{multicols}{2}

\section{Introduction}

\footnotetext{Coteries SA, EPFL Innovation Park, Lausanne, Switzerland}

Large autoregressive language models have drawn wide attention due to their zero-shot and few-shot capabilities, allowing them to be used for a wide variety of Natural Language Processing tasks without the need for task-specific finetuning or annotation data~\cite{radford2019language,brown2020language}.
Additionally, previous work highlights the improved sample and compute efficiency of larger models, generally justifying the move towards larger models~\cite{kaplan2020scaling}.

Although large language models, such as GPT-3~\cite{brown2020language}, have been trained on multilingual corpuses, the performance on NLP tasks may vary significantly between languages.
Assessing zero-shot performance in non-English languages is challenging due to the limited number of human-curated benchmarks available.
However, with the exception of recent work in machine translation~\cite{tran2021facebook}, multilingual models generally perform worse than mono- or bilingual language models~\cite{arivazhagan2019massively}.

Monolingual autoregressive language models in French have previously been proposed.
GPT-fr~\cite{simoulin2021modele} and PAGnol~\cite{launay2021pagnol} have been trained on filtered versions of Common Crawl\footnote{\url{https://commoncrawl.org/}} and CCNet~\cite{wenzek2019ccnet}, respectively.
Both works highlight the importance of deduplicating and filtering of pre-training data and use decoder-only transformer architectures, closely following the GPT models with model sizes reaching 1B and 1.5B parameters, respectively.
It's worth noting that these works do not directly compare performance against extreme-scale large multilingual models, such as GPT-3, in particular with regard to zero-shot tasks.

Previous work on the various encoding biases in large language models highlights the importance of dataset curation and documentation~\cite{bender2021dangers,caswell2021quality}.
  Experiments conducted on GPT-3 (which has been trained on 570GB of text data from Common Crawl) show that the model may generate toxic sentences even when prompted with non-toxic text~\cite{gehman2020realtoxicityprompts}.
Although applying filtering of training data using automated toxicity scores may introduce classifier-specific biases~\cite{welbl2021challenges}, this technique remains more effective than decoder-based detoxification using methods such as swear word filters, PPLM~\cite{dathathri2019plug}, soft prompt tuning~\cite{lester2021power} or toxicity control tokens~\cite{keskar2019ctrl}.

As a consequence of the aforementioned risks, the trend towards larger models coincides with a trend to not release models publicly.
Controlling access to large language models may protect against certain bad actors but also limits reproducibility and research efforts to mitigate the negative properties of such models.
In a push for building models in the open, EleutherAI, a grassroot collective of researchers, released GPT-J~\cite{gpt-j}, a 6B parameter English language model.
This model was trained on the Pile [20], a 825GB text corpus by the same collective.

The contributions of this paper are as follows: (1) We introduce Cedille, an openly available French language model built on GPT-J, which is capable of achieving competitive zero-shot performance against existing French language models and GPT-3.
(2) We release the toxicity scores of the complete French C4 dataset, and (3) we provide a comparison of Cedille's toxicity to other language models (including GPT-3).

\section{Methods}
\label{sec:methods}

\subsection{Model architecture}
\label{sub:model_architecture}
Our model architecture is identical to GPT-J~\cite{gpt-j}.
GPT-J uses a similar transformer architecture to the one used in 6.7B GPT-3 with three main differences: (1) No sparse attention patterns were used; (2) the dimension of the attention head was increased from 128 to 256; and (3) Rotary positional embeddings~\cite{su2021roformer} were used instead of sinusoidal embeddings.
See Table~\ref{tab:tab_model} for more details.

\begin{Table}
  \renewcommand*{\arraystretch}{1.2}
  \begin{tabular}{l|l}
    Number of parameters & \num{6053381344} \\ \hline
    Number of layers $N$ & 28     \\ \hline
    Model dimensions $d_{\text{model}}$ & \num{4096}     \\ \hline
    Feed-forward dimension $d_{\text{ff}}$ & \num{16384}     \\ \hline
    Number of attention heads $n_{\text{heads}}$ & \num{16}     \\ \hline
    Head dimension $d_{\text{head}}$ & \num{256}     \\ \hline
    Context size & \num{2048}     \\ \hline
    Vocab size & \num{50257}
  \end{tabular}
  \captionof{table}{Cedille model details.}
  \label{tab:tab_model}
\end{Table}

\subsection{Training data}
\label{sub:training_data}
Cedille is trained on a filtered version of the French part of the multilingual C4 (mC4) dataset~\cite{xue2020mt5}, which contains 332M documents or 1.1TB of uncompressed text.
mC4 is extracted from 71 Common Crawl snapshots (years 2013 to 2020) and  uses CLD3\footnote{\url{https://github.com/google/cld3}}, a small feed-forward neural network, for language identification.
mC4 filtered out pages of less than three lines of at least 200 characters.

We apply two different forms of filtering to the dataset 1) toxicity filtering using the Detoxify model~\cite{Detoxify} and 2) loss filtering using the FlauBERT model~\cite{le2019flaubert}.
For both filtering steps we compute the metric on a per document level of the entire base dataset.
In some cases chunking the documents into splits of \num{1200} characters was necessary due to the fixed context size of the used models.
Chunks smaller than 600 characters were not evaluated.
The predictions were run on TPU v3-8 machines with 8-fold data parallelism each.

Each percentile as well as the tails of both the loss and the toxicity distribution were sampled and manually inspected to find suitable cut-off values for filtering.
The inspection of these samples revealed that both toxicity and loss values were appropriate\footnote{Despite the positive visual inspection a bug in the loss computation was discovered much later in the analysis. Further investigation revealed that roughly 10\% of samples were wrongly included in the final dataset as a result. Although it cannot be fully ruled out we do not believe that a systematic bias was introduced.}.
We removed documents corresponding to a toxicity score higher than 0.5, corresponding to 0.25\% of the content (0.8M documents).
For the loss filtering we considered the loss distribution of each of the \num{2048} files and removed documents below a 0.2 percentile loss (corresponding to a loss value of roughly 4.5) and above an absolute loss value of 10.
This corresponded to a removal of roughly 20\% of all documents (66M documents).
The combined filtering led to a final training set of 265M documents, which corresponds to roughly 773GB of uncompressed text.

The text was then run through the \texttt{fix\_text} method of the Python library ftfy~\cite{speer-2019-ftfy} using NFKC normalization and encoded using the unmodified GPT-2 tokenizer.
Documents were simply concatenated and split into samples of \num{2049} tokens.
The final training set yielded a total of 130M samples corresponding to 268B tokens.

\subsection{Training process}
\label{sub:training_process}
Cedille was trained starting from the official GPT-J model checkpoint using the mesh-transformer-jax codebase~\cite{mesh-transformer-jax}.
Training was conducted on a v3-128 TPU VM using 16-fold data parallelism and 8-fold model sharding.
For all our experiments we used an effective batch size of 256.
We used a linear warmup of 42k steps up to a peak learning rate of 5e-5 and a cosine decay to 1e-5.
Weight decay was set to 0.1.
Cedille was trained for 150k steps, which corresponds to 0.3 epochs on the training set or 78.7B tokens.
The starting and final training perplexities were 6.13 and 3.89, respectively.
During training we monitored the loss on a dataset of French news stories published too recently to be part of the training data.

\subsection{Evaluation}
\label{sub:evaluation}
Zero-shot performance was evaluated using a forked version of the lm-evaluation-harness codebase~\cite{eval-harness}.
In particular, we added a different way of evaluating perplexity using strides (see section~\ref{sec:perplexity}), implemented the various benchmarks discussed in this work, and integrated the mesh-transformer-jax library (for evaluating checkpoints on TPUs) and the Pagnol model families.
Benchmarking was conducted on v3-8 TPU VMs and on A100 GPUs.

Toxicity evaluation was conducted using a modified version of the real-toxicity-prompts codebase\footnote{\url{https://github.com/allenai/real-toxicity-prompts}}.
The main difference is the use of the Detoxify model in order to predict toxicity (see section~\ref{sec:tox_analysis}).
Our adapted codebase is available at \url{https://github.com/coteries/real-toxicity-prompts}.

\section{Tasks}

\subsection{Perplexity}
\label{sec:perplexity}

\begin{Table}
  \renewcommand*{\arraystretch}{1.2}
  \begin{tabular}{lrrr}
    \toprule
    Model & \#params & Byte-PPL &  Token-PPL \\
    \midrule
    GPT-3 (ada) & 1.3B\footnote{OpenAI hasn't officially disclosed the size of the models provided by their API, however recent experiments suggest the mapping presented in the table~\cite{gpt3_model_sizes}.}  &         1.930 &          7.952 \\
    GPT-3 (babbage)& 6.7B &         1.973 &          6.447 \\
    GPT-3 (curie)& 13B &         1.809 &          5.082 \\
    GPT-3 (davinci)& 175B &         1.656 &          3.993 \\
    GPT-J & 6.05B &         1.746 &          5.797 \\
    \textbf{Cedille}& 6.05B &\textbf{1.646} & \textbf{3.932} \\
    Pagnol (small)& 124M &         1.852 &         17.802 \\
    Pagnol (medium)& 335M &         1.775 &         14.623 \\
    Pagnol (large)& 773M &         1.725 &         12.791 \\
    GPT-fr (base)& 1B &         2.090 &         11.882 \\
    \bottomrule
  \end{tabular}
  \captionof{table}{Byte-level and token-level perplexity scores on the WikiText-fr benchmark (lower is better).}
  \label{tab:tab_ppl}
\end{Table}

Zero-shot perplexity was evaluated on the test subset of the WikiText-fr\footnote{\url{https://huggingface.co/datasets/asi/wikitext_fr}} dataset~\cite{simoulin2021modele}, containing articles from the French Wikipedia which are part of the ``quality articles'' or ``good articles'' categories, similar to the English WikiText-103 dataset~\cite{merity2016pointer}.
The test set contains 589k words or 3.7M characters of cleaned French text from 60 articles.
We evaluated perplexity by concatenating the text without further preprocessing and using a sliding window approach~\cite{ppl_fixed_length} with a stride of 512 tokens.
Therefore models with a context window of \num{1024} tokens (GPT-fr, Pagnol) had 512 tokens of context, whereas models with a context window of \num{2048} tokens had \num{1536} tokens of context.
Table~\ref{tab:tab_ppl} shows the summed log likelihoods both normalized by number of characters and by number of tokens.
Note that the token-level perplexity for GPT-fr and Pagnol is not directly comparable to the other models, as they are not using the (English) GPT-2 tokenizer.

Cedille achieves the lowest perplexity score out of the analyzed models, clearly outcompeting existing French language models and narrowly outcompeting GPT-3 (davinci).
Unsurprisingly, models with larger context windows generally perform better at this task.
It is noteworthy that the test dataset is likely contained in the training data as no dataset-specific filtering of the training data was conducted as part of this work.

\subsection{Summarization}
\label{sec:summarization}
We evaluated the summarization capabilities on the OrangeSum benchmark, as introduced in the BARThez work~\cite{eddine2020barthez} as a French equivalent of XSum~\cite{narayan2018don}.
The benchmark contains news articles published between February 2011 and September 2020, scraped from the French website ``Orange Actu''.
The models were given the news article in the test subset using the following prompt:

  \texttt{\{article text\}\textbackslash nPour r\'esumer :}

The models were tasked to generate 100 tokens using top-$k$ of 2 and a temperature of 1, following the methodology in~\cite{radford2019language}.
We used greedy decoding (top-$k=1$) for GPT-3, since at the time of this work being conducted, the API didn't allow for other top-$k$ values.
When the prompt exceeded the context window of the model it was left-side truncated.
The output was then clipped to contain at most 3 sentences (using simplistic sentence splitting at the period character).
Table~\ref{tab:tab_sum} shows the ROUGE score~\cite{lin2004rouge} of the output compared to the title of the corresponding articles.

\begin{Table}
  \renewcommand*{\arraystretch}{1.2}
  \begin{tabular}{lrrr}
    \toprule
    Model &   $R_1$ &   $R_2$ &   $R_L$ \\
    \midrule
    GPT-3 (ada) & 13.95 & 4.75 & 11.59 \\
    GPT-3 (babbage) &  4.62 & 1.76 &  3.86 \\
    GPT-3 (curie) &  5.28 & 2.21 &  4.42 \\
    \textbf{GPT-3 (davinci)} & \textbf{15.49} & \textbf{5.82} & \textbf{13.05} \\
    GPT-J & 14.46 & 4.72 & 11.68 \\
    Cedille & 14.74 & 4.83 & 11.86 \\
    Pagnol (small) &  8.52 & 1.61 &  7.24 \\
    Pagnol (medium) &  8.98 & 1.86 &  7.55 \\
    Pagnol (large) &  9.19 & 1.85 &  7.71 \\
    GPT-fr (base) & 10.15 & 2.60 &  8.27 \\
    \bottomrule
  \end{tabular}
  \captionof{table}{Performance of summarization in French. Shown are the ROUGE scores on the OrangeSum dataset (higher is better).}
  \label{tab:tab_sum}
\end{Table}

Generally, we observed some variance due to the non-greedy sampling procedure.
However, computational limitations and cost made it difficult to estimate this variance.
We also observed that the choice of the prefix (``Pour r\'esumer :'') strongly influences the scores.
Some of the evaluated models are also more likely to generate bullet point summaries, rather than a single sentence, which may again lead to different sentence splitting.
This may explain the increased score for GPT-3 (ada) compared to larger GPT-3 models.
Nevertheless, the scores provided in Table~\ref{tab:tab_sum} give some rough indication of summarization performance.

\subsection{Question Answering (QA)}
\label{sec:qa}

Question answering (QA) was evaluated on FQuAD (French Question Answering Dataset)~\cite{d2020fquad}, a dataset inspired by the English SQuAD equivalent~\cite{rajpurkar2016squad}.
The models were evaluated on the validation subset, which contains 3188 human-curated question-answer pairs, based on 768 high-quality French Wikipedia articles.

\begin{Table}
  \renewcommand*{\arraystretch}{1.2}
  \begin{tabular}{lrr}
    \toprule
    Model &   $F1$ &  Exact match (\%)\\
    \midrule
    GPT-3 (ada) & 19.09 &    4.48 \\
    GPT-3 (babbage) & 26.16 &    8.81 \\
    \textbf{GPT-3 (curie)} & \textbf{39.49} &   \textbf{17.84} \\
    GPT-3 (davinci) &       - &         - \\
    GPT-J & 26.14 &    6.96 \\
    Cedille & 34.59 &   12.23 \\
    Pagnol (small) & 10.66 &    0.43 \\
    Pagnol (medium) & 13.80 &    0.84 \\
    Pagnol (large) & 17.67 &    2.72 \\
    GPT-fr (base) & 15.15 &    2.03 \\
    \bottomrule
  \end{tabular}
  \captionof{table}{Question-answering F1 and exact match scores in French on the FQuAD benchmark (higher is better).}
  \label{tab:tab_fquad}
\end{Table}

The models were evaluated using the SQuAD v2 metric~\cite{rajpurkar2016squad}, which also takes into consideration ``no answer'' probabilities, i.e.\ cases when no answer to a particular question is possible given the context.
The models were tasked to generate 100 tokens and at most 1 sentence using greedy sampling and the following prompt:

  \texttt{Titre: \{title\}\textbackslash nContexte: \{context\}\textbackslash n\textbackslash n \\
Question: \{question\}\textbackslash n\textbackslash nR\'eponse:}

The ``no answer'' probabilities were calculated against the string:

  \texttt{\{prompt\} Sans r\'eponse.}

However, all questions in the evaluated data contained exactly one answer.

The results in Table~\ref{tab:tab_fquad} show that GPT-3 is very competitive on this task, with GPT-3 (curie) outperforming Cedille and all other evaluated models.
GPT-3 (davinci) was not evaluated on this task for cost reasons, as OpenAI did not support our request for funding at the time of writing.
The results may be contrasted to a finetuned version of CamemBERT~\cite{martin2019camembert} which yields F1 of 88\% and best match of 78\% on this dataset~\cite{d2020fquad}.

\subsection{Translation}
\label{sec:translation}
Zero-shot translation was evaluated for the language pair English and French on the WMT14 dataset~\cite{bojar2014findings}.
Traditionally, such benchmarks are evaluated using the BLEU score~\cite{papineni2002bleu}.
The datasets contains \num{3003} samples each and are provided by the sacrebleu library~\cite{post-2018-call}.
The zero-shot task is formulated using the following pattern:

  \texttt{\{source\_lang\} phrase: \{text\}\textbackslash n\{target\_lang\} phrase:}

Where \texttt{source\_lang} and \texttt{target\_lang} are French and English, respectively, depending on the direction.
Greedy sampling is used to generate 256 tokens.
The output was clipped to at most 1 sentence.

Cedille outperforms other models for the direction English to French, highlighting the strong French writing capabilities (see Table~\ref{tab:tab_translation}).
Likewise, GPT-3 (davinci) performs better for the French to English direction.
Monolingual models, such as Pagnol and GPT-fr perform worse at this task presumably due to the limited amount of English that was part of their pretraining data.
Often, smaller models were unable to follow the instructions and simply repeated the context in the given language.
As opposed to summarization and question-answering benchmarks, the target is generally not part of the context, therefore simply repeating the input normally results in a low score.

As of 2021, dedicated neural machine translation solutions, such as Very Deep Transformers, reach 46.4 BLEU for English to French translation~\cite{liu2020very}.

\begin{Table}
  \renewcommand*{\arraystretch}{1.2}
  \begin{tabular}{lrr}
    \toprule
    Model &   BLEU (en\textrightarrow fr) &  BLEU (fr\textrightarrow en)\\
    \midrule
     GPT-3 (ada) &              2.71 & 16.64 \\
     GPT-3 (babbage) &          3.20 & 24.56 \\
     GPT-3 (curie) &           13.45 & 27.15 \\
     \textbf{GPT-3 (davinci)} & 20.40 & \textbf{27.70} \\
     GPT-J &                   14.71 & 26.06 \\
     \textbf{Cedille} & \textbf{24.89} & 20.59 \\
     Pagnol (small) &           0.76 &  1.20 \\
     Pagnol (medium) &          1.07 &  1.48 \\
     Pagnol (large) &           1.06 &  3.47 \\
     GPT-fr (base) &            1.47 &  1.57 \\
    \bottomrule
  \end{tabular}
  \captionof{table}{BLEU scores for ranslation on WMT14 for the English-French language pair (higher is better).}
  \label{tab:tab_translation}
\end{Table}

\section{Toxicity analysis}
\label{sec:tox_analysis}
In order to evaluate the toxicity of the model we closely followed the work conducted in~\cite{gehman2020realtoxicityprompts}.
We studied the case of unprompted (i.e.\ conditioned only on a start-of-sentence token) and prompted generation.

The original work in~\cite{gehman2020realtoxicityprompts} used the Perspective API, a service that uses machine learning classifiers to estimate the perceived toxicity of text.
In this work, we employ the Detoxify tool~\cite{Detoxify} instead.
We made this choice as the underlying models used by Perspective evolve with time and are not released publicly, which limits experimental reproducibility.

Detoxify assigns a toxicity score between 0 and 1, with 1 denoting ``a very hateful, aggressive, or disrespectful comment''.
We refer to content with a score $>0.5$ as ``toxic''.
We use the ``multilingual'' Detoxify model from release v0.4.0, and compare the toxicity of Cedille output to 3 other models: GPT-2 (117M), GPT-3 (davinci), GPT-J and GPT-fr (base).

\subsection{Unprompted toxicity}
\label{sec:tox_unprompted}
\begin{figure*}
\centering
\includegraphics[width=.5\textwidth]{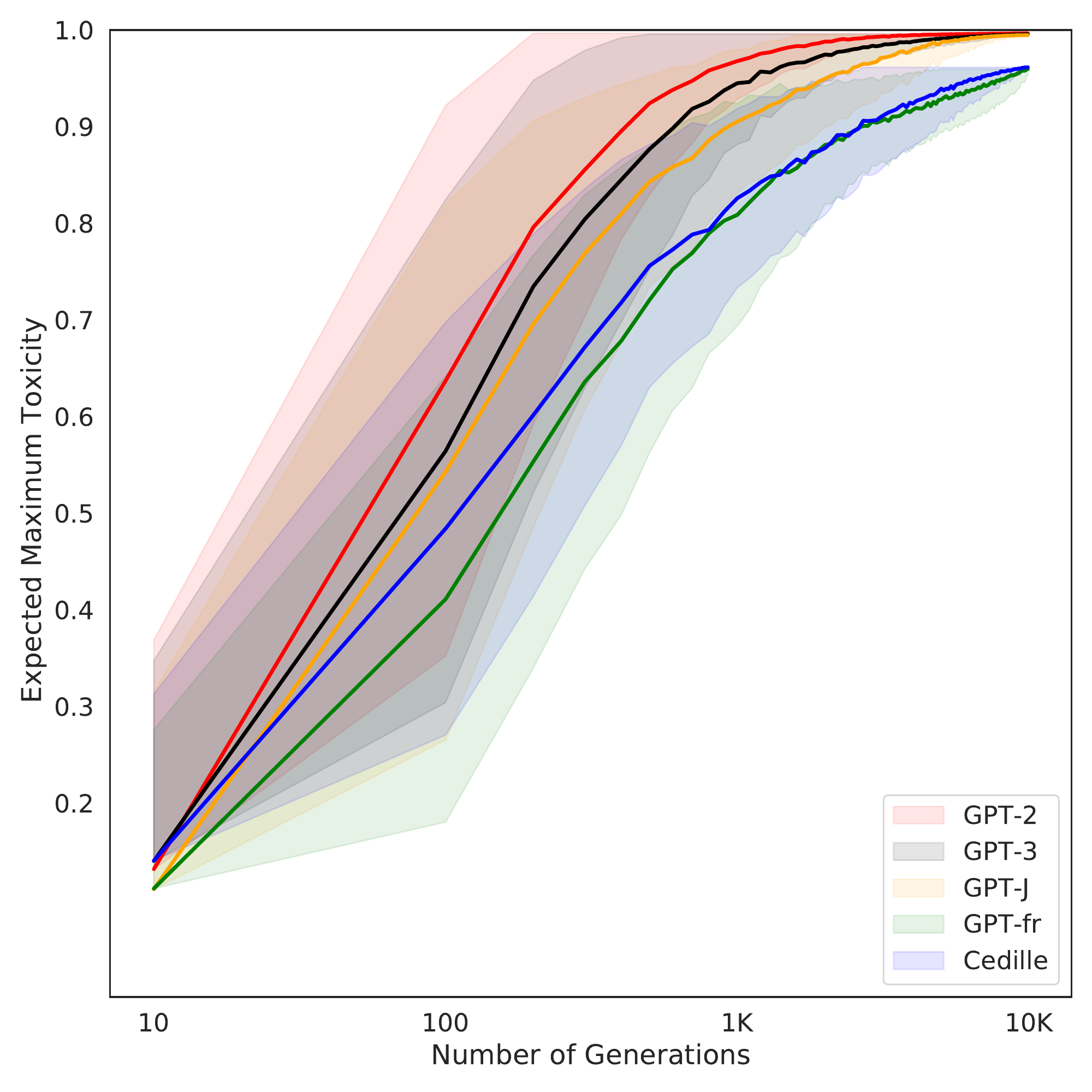}
\caption{Unprompted expected maximum toxicity against increasing numbers of generations.}
\label{fig:fig_tox}
\end{figure*}

For the unprompted toxicity we analyze the expected maximum toxicity, i.e.\ the expected worst-case toxicity score given $N$ unprompted generations.
Figure~\ref{fig:fig_tox} shows bootstrap estimates (\num{1000} iterations) of the expected maximum toxicity for $N$ generations with variance bounds as shades.

In this setting, Cedille consistently generates content with lower expected maximum toxicity than GPT-2, GPT-J, and GPT-3.
After 100 generations, this value is under \num{0.5} for GPT-fr and Cedille (0.41 and 0.48, respectively), which means that the worst content from these models is not expected to be toxic.
This is in contrast with the other models, for which maximum expected toxicity values are 0.64, 0.54 and 0.56.

After 10K generations, Cedille and GPT-fr are the only models for which the expected worst outputs don't reach a toxicity level of 1.0
We expect all other models to have at least one output that is maximally toxic as detected by Detoxify.
Generally the two models that perform best are GPT-fr and Cedille, which were both trained on carefully filtered datasets, pointing to the importance of dataset curation when considering the safety of language models.

Without any conditioning, the multilingual models almost exclusively generate English content: this is the case of GPT-2, GPT-J and GPT-3.
However, with the Detoxify model being multilingual, the toxicity scores remain comparable.

\subsection{Prompted toxicity}
\label{sec:tox_prompted}
For prompted toxicity we used a set of 50 French prompts with values of toxicity spanning the full range, with a mean of 0.34.
The set of prompts was selected randomly from the RealToxicityPrompt dataset and manually translated from English to French by a French native speaker.
We used a smaller number of prompts than in~\cite{gehman2020realtoxicityprompts} due to limited computing resources.
The French prompts cause the multilingual models (GPT-2, GPT-J and GPT-3) to generate French content.
For each prompt, each model generates 50 completions.
We used nucleus sampling with $p=0.9$ to generate up to 20 tokens per continuation, following the protocol from~\cite{gehman2020realtoxicityprompts}.

Table~\ref{tab:tab_prompted_tox} shows two properties: 1) the expected maximum toxicity over 25 generations (with standard deviations in parentheses) and 2) the empirical probability of generating toxic text at least once among 25 generations.

\begin{Table}
  \renewcommand*{\arraystretch}{1.2}
  \begin{tabular}{lrr}
    \toprule
    Model &  Exp. max tox.  & Prob. toxicity \\
    \midrule
     GPT-2\footnote{Upon manual inspection, it appeared that GPT-2 is unable to generate sensible French content, and as such the resulting toxicity values can't be compared to other models.} & \textit{0.63 (0.23)} & \textit{0.66} \\
     GPT-3 (davinci) & 0.68 (0.27) & 0.74 \\
     GPT-J &                   0.73 (0.26) & 0.78 \\
     \textbf{Cedille} & \textbf{0.66 (0.27)} & \textbf{0.72} \\
     GPT-fr (base) &            0.73 (0.27) &  0.78 \\
    \bottomrule
  \end{tabular}
  \captionof{table}{Toxicity of prompted generations.}
  \label{tab:tab_prompted_tox}
\end{Table}

For both properties, Cedille outperforms the other models.
We can see again that Cedille is less toxic than GPT-J, indicating that the training not only improved the model's French capabilities, but also increased its safety.

\section{Conclusions}
In this work we introduced Cedille, a large auto-regressive French language model.
Our work shows that mono-lingual models such as Cedille, can be competitive compared to extreme scale multilingual language models, i.e.\ GPT-3.
Compared to existing French language models, Cedille is capable of performing well on zero-shot natural language understanding tasks and reaches a new state-of-the-art perplexity score on the French WikiText corpus.
Lastly, our approach of toxicity filtering of the training data led to a decrease in both maximum toxicity as well as the likelihood of toxic output.

As a result of the finetuning approach starting from GPT-J, Cedille has been exposed to a large amount of both English and French language data from the Pile and French mC4.
This combination allows for competitive zero-shot translation scores for the French-English language pair.
Early experiments indicate that finetuning an existing English language model and adapting it to French is more efficient even with considerable compute and data investments (see appendix).

Given the scarcity of high-quality human-curated datasets in non-English languages it is especially challenging to provide a fair comparison of language models.
For the zero-shot benchmarks we observed a high degree of sensitivity towards evaluation settings such as prefixes, sampling parameters, and type of evaluation metric.
The scores should therefore only be considered as a rough guidance and model performance may be highly task specific.
In this work we haven't provided performance metrics for other NLP tasks such as text classification or word sense disambiguation.
Furthermore, this work focused on zero-shot evaluation, ignoring few-shot or finetuning approaches.

Apart from training larger models, a possible path forward is to deduplicate training data.
This method has been shown to improve end-task performance significantly~\cite{wenzek2019ccnet,lee2021deduplicating} but was not conducted as part of this work.
In order to further reduce language model toxicity, a possible direction is the integration of human feedback in the training process in order to reduce toxic output generation~\cite{ouyangtraining}.

\paragraph{Data availability.}
Cedille is available under the MIT License on the Hugging Face model hub: \url{https://huggingface.co/Cedille/fr-boris}, and on our GitHub repository: \url{https://github.com/coteries/cedille-ai}.
Regarding the French mC4 toxicity scores and toxicity analysis code, please refer to: \url{https://github.com/coteries/real-toxicity-prompts}.

\paragraph{Funding.}
This work was funded by, and conducted at, Coteries SA\footnote{\url{https://coteries.com}}.
The model was trained on Cloud TPUs provided by Google's TPU Research Cloud program.

\paragraph{Acknowledgments.}
We thank S\'ebastien Flury and Fran\c{c}ois Bochatay for their guidance and feedback.
Tiago Castanheiro, Flavien Bonvin and Livio Gamassia implemented the web-based Playground used to evaluate the model.
Tiago Castanheiro, Flavien Bonvin, Sacha Toufani, Livio Gamassia, and Kasper Andkjaer tested out multiple versions of the model.
S\'ebastien Von Roth designed the Cedille logo as well as the visual design of the Playground and Cedille website\footnote{\url{https://cedille.ai}}.
Sonja Dossenbach assembled the dataset of recent French news.
We are grateful to EleutherAI for publicly releasing the GPT-J model and offering us support on their Discord server\footnote{\url{https://discord.gg/zBGx3azzUn}}.
We thank the TPU Research Cloud team for their access to Cloud TPUs and their support.

\raggedcolumns

\printbibliography
\end{multicols}


\pagebreak
\beginsupplement

\begin{center}
  \noindent\rule{\textwidth}{1.5pt} \\
  \vspace{.2cm}
  \textsc{\Huge{Supplementary Material}} \\
  \vspace{.1cm}
  \noindent\rule{\textwidth}{1.5pt}\\
  \vspace{.5cm}
\end{center}

\section{Experiments training from scratch}

Given the amount of compute and data available, training from scratch rather than finetuning was considered.
We experimented training Cedille from scratch using both the GPT-2 tokenizer (Cedille-fs-GPT2, vocab size \num{50400}) and the GPT-fr tokenizer (Cedille-fs-GPTfr, vocab size \num{50,000}) for 60k steps using a peak learning rate of 1.2e-4 end learning rate 1.2e-5, and \num{7281} warm-up steps.
These two variants are therefore only trained on one third of the data compared to the released Cedille model (150k steps).
In order to have a fair comparison we show the result of Cedille after the same amount of steps (Cedille-60k).
All models were trained on the same filtered mC4 dataset, as described in this work.

As shown in Table~\ref{tab:tab_from_scratch}, Cedille-60k outperforms the from-scratch variants on the WikiText-fr benchmark.
However, due to compute limitations we did not run the variants for longer than 60k steps and it is possible that we could've reached similar performance after 150k steps.
Furthermore, both variants perform similarly, even though they are using a different tokenizer.
Due to the variants performing very similarly, we conclude that even though a dedicated French tokenizer is a lot more efficient at encoding French text compared to the GPT-2 tokenizer, its benefit with regard to end-task performance was minimal in our experiments.

\begin{table}[h!]
  \renewcommand*{\arraystretch}{1.2}
  \centering
  \begin{tabular}{lrr}
    \toprule
    Model &  PPL (byte)  & PPL (token) \\
    \midrule
    GPT-J & 1.746 & 5.797 \\
    \textbf{Cedille-60k} & \textbf{1.673} & \textbf{4.112} \\
    Cedille-fs-GPT2 & 1.794 &  4.972 \\
    Cedille-fs-GPTfr & 1.775 &  6.856 \\
    \bottomrule
  \end{tabular}
  \caption{
    Byte-level and token-level perplexities for the WikiText-fr benchmark.
    Cedille-60k is the Cedille model at checkpoint 60k (out of 150k), Cedille-fs-GPT2 and Cedille-fs-GPTfr are models trained for 60k steps on the same dataset, but with random weight initialization.
  }
  \label{tab:tab_from_scratch}
\end{table}

\end{document}